\documentclass[10pt,twocolumn,letterpaper]{article}

\usepackage{iccv}
\usepackage{times}
\usepackage{epsfig}
\usepackage{graphicx}
\usepackage{amsmath}
\usepackage{amssymb}

\usepackage{booktabs}
\usepackage{subcaption}
\usepackage{float}
\usepackage{multirow}
\usepackage{multicol}
\usepackage{makecell}
\usepackage{pifont}
\usepackage{bbm}

\usepackage[accsupp]{axessibility}
\usepackage{amsmath}  

\usepackage[
breaklinks=true,
bookmarks=true,
colorlinks=true,
pdfpagemode=UseOutlines, 
bookmarksnumbered=true   
]{hyperref}

\iccvfinalcopy 

\def\iccvPaperID{****} 

\usepackage[bottom]{footmisc}
\begin{document}

\title{Transformer-Based Dual-Optical Attention Fusion Crowd Head Point Counting and Localization Network}

\author{Fei Zhou$^{1,2}$ \quad Yi Li$^1$\thanks{corresponding author} \quad Mingqing Zhu$^2$\\
		$^1$
        Neusoft Institute Guangdong, China $^2$Airace Technology Co.,Ltd., China\\
		{\tt\small \{zhoufei21, liyi\}@s.nuit.edu.cn}
	}

\maketitle


\begin{abstract}

In this paper, the dual-optical attention fusion crowd head point counting model (TAPNet) is proposed to address the problem of the difficulty of accurate counting in complex scenes such as crowd dense occlusion and low light in crowd counting tasks under UAV view. The model designs a dual-optical attention fusion module (DAFP) by introducing complementary information from infrared images to improve the accuracy and robustness of all-day crowd counting. In order to fully utilize different modal information and solve the problem of inaccurate localization caused by systematic misalignment between image pairs, this paper also proposes an adaptive two-optical feature decomposition fusion module (AFDF). In addition, we optimize the training strategy to improve the model robustness through spatial random offset data augmentation. Experiments on two challenging public datasets, DroneRGBT and GAIIC2, show that the proposed method outperforms existing techniques in terms of performance, especially in challenging dense low-light scenes. Code is available at \url{https://github.com/zz-zik/TAPNet}.

\end{abstract}


\begin{figure}[ht]
	\centering
	\includegraphics[width=1.0\linewidth]{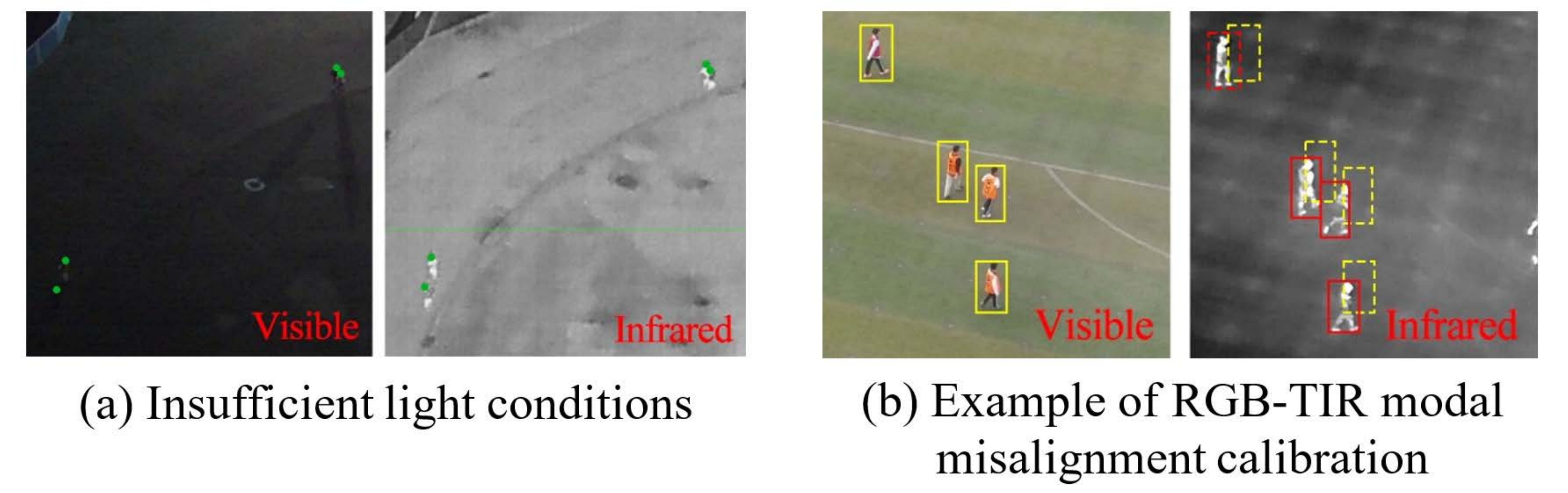}
	\caption{\textbf{Examples of infrared and visible images}. (a) The two rows of people on the left are almost invisible in the visible spectrum under low light conditions, illustrating the fact that IR images are more advantageous in low light conditions. (b) Example of RGB-TIR modal misalignment, showing that the modal misalignment problem is more prominent in target detection from the UAV viewpoint, where the yellow and red boxes denote the annotations of the same objects in the TIR image and the RGB image, respectively.
	}
	\vspace{-10pt} 
	\label{fig:fig1_concept}
\end{figure}

\section{Introduction}

The crowd counting task aims to count the number of people in visual content such as images and videos. The technique plays an important role in many scenarios including urban traffic management, mall traffic analysis and large event crowd monitoring~\cite{jiang2020attention,liu2022leveraging,wan2021generalized}. However, traditional visible-light based crowd counting methods are limited by imaging constraints under adverse conditions such as nighttime, and cannot fully perceive the target. As shown in (a) in Fig.~\ref{fig:fig1_concept}a, visible spectrum-based object detection may lack information, leading to missed or false alarms. Multi-spectral information combines complementary information from between different modalities and can improve the perception, reliability and robustness of the detection algorithm. Therefore, fusion of different imaging modalities for multimodal imaging perception can achieve complementary information from multimodal images, greatly enhancing the ability of multidimensional high-resolution observation, and perceiving the physical world in a more comprehensive, clearer, and more accurate way.

The problem of modal misalignment is still faced in multispectral object detection, and most feature fusion methods usually assume that the RGB-TIR images are well aligned.Yuan et al~\cite{yuan2023c2former} showed that RGB-TIR image pairs are captured by sensors with different fields of view (FoVs) at different imaging timestamps. As a result, imaging objects captured in both modalities usually suffer from misalignment problems. As shown in (b) in Fig.~\ref{fig:fig1_concept}b, the modal misalignment problem is more prominent in target detection from the UAV viewpoint, as targets are usually labelled using tightly oriented bounding boxes.  And the dual challenge of weak spatial alignment superimposed on small targets leads to the poor performance of common multimodal fusion methods, making the design of fusion strategies extremely challenging. Therefore, how to effectively fuse feature representations between different light sources, make full use of the intrinsic complementarity between different modalities, and design effective cross-modal fusion mechanisms in order to obtain the maximum performance gain? Thus, the accuracy and robustness of the model for crowd counting in the many complex scenarios mentioned above can be improved.

In this work, we formulate the crowd counting task as a dual-light fusion head-point matching process. Specifically, the method cleverly fuses feature representations between different light sources through a dual-light attentional feature fusion module, combined with a head-point matching network with auxiliary point guidance, to further improve the model's counting capability in complex scenes. In contrast, the point-based approach has the advantage of using learnable point matching to directly use point labelling as the learning target, which simplifies the process of localisation and improves the detection efficiency of the model by guiding the regression of individual point coordinates. The bimodal fusion method is able to capture the features of the target individual more comprehensively and reduce the misjudgement due to light changes or background interference, thus improving the reliability of model counting.

In order to solve the problem of inaccurate localisation caused by systematic misalignment between image pairs, this paper proposes a new adaptive two-branch feature decomposition fusion module (AFDF), which is able to effectively align potential spatial features between modalities. More specifically, AFDF can perform both intra-modal and inter-modal fusion and robustly capture potential interactions between RGB and TIR by exploiting the Transformer's self-attention mechanism.

Extensive experimental results show that our approach can significantly improve the accuracy and reliability of crowd counting models. The contributions of the work in this paper are mainly in the following three areas:

\begin{itemize}
    \setlength\itemsep{0.1em}
    \item By introducing the complementary information of infrared images, we propose a bi-optical attention fusion crowd head point counting model, thus compensating for the model's counting limitation under adverse conditions such as nighttime.     
    
    \item In order to solve the problem of inaccurate localisation caused by systematic misalignment between image pairs, an adaptive two-branch feature decomposition fusion module (AFDF) is proposed.
    
    \item We also optimise the training strategy, i.e., the spatial random offset data enhancement strategy, to further improve the overall accuracy of the model in point localisation and the robustness of point matching.
\end{itemize}


\section{Related Work} 

We briefly divide the existing work based on the crowd counting methods used, i.e., detection based methods, density map based methods and point localisation based methods. And we also discuss the latest advances in multispectral image fusion and multispectral modal mismatch fusion.

\subsection{Relevant methods for population counting} 

\vspace{-5pt} 
\paragraph{Detection-based approach.}
If is implemented based on Faster RCNN~\cite{ren2015faster}. Specifically, LSC-CNN~\cite{sam2019locate} employs a multicolumn architecture and a top-down feedback processing mechanism, which uses headpoint features to generate pseudo bounding boxes to estimate the number of people in an image.PSDDN~\cite{liu2019point} proposes to initialise pseudo bounding boxes based on nearest-neighbour distances, and introduces an on-line updating scheme to optimise the training process, from which smaller prediction frames are selected to update the pseudo frames as a way of increasing the detection accuracy.The YOLO family of algorithms stands out for its concise and clear structure as well as its wide range of applications.DroneNet~\cite{wang2023dronenet} uses YOLOv5 as a backbone network and proposes a split-concat feature pyramid network (SCFPN) for fusing feature information from different scales. Although these methods have achieved good results, they all ignore the problem of inconsistent head point features caused by multi-scale variations in sparse scenes.

\vspace{-10pt}
\paragraph{Methods based on density maps.}
is a common method for most crowd counting tasks and it was first introduced in ~\cite{wang2023dronenet}. The core idea of this method is to map the crowd density to each pixel of the image, thus generating a density map that is able to predict the number of people directly from low-level features by summing the predicted density maps. Therefore this method first requires the use of a Gaussian kernel to generate the ground-truth density map used as labels before network training, and the Gaussian kernel is capable of generating smooth density distributions based on the location of a person's head or body parts. In recent years, many cutting-edge works have been devoted to advancing the counting performance of such methods.Idress et al ~\cite{idrees2018composition} used a small Gaussian kernel to generate density maps, and although using a small kernel generates clear density maps, it still fails to address the problem of overlapping in extremely dense regions. To address this problem, several approaches ~\cite{lin2023optimal,liang2021focal,gao2020learning} focus on designing new density maps such as distance labelled density maps ~\cite{lin2023optimal}, Focused Inverse Distance Transformed Density Maps (FIDTM) ~\cite{liang2021focal} and Independent Instance Density Maps (IIM) ~\cite{gao2020learning}. Although these methods have made significant progress in counting performance, they still suffer from some inherent shortcomings, such as the inability to provide information about the exact location of an individual in a population and a significant dependence on the quality of the density map.

\vspace{-10pt}
\paragraph{Point-based positioning methods.}
Song et al ~\cite{song2021rethinking} first proposed a purely point-based joint framework for crowd counting and individual localisation called peer-to-peer network (P2PNet) in 2021. The method provides a fine-grained solution in the field of crowd counting. Specifically, the point-based approach estimates the number and location of the crowd mainly by identifying and locating individuals (usually heads or body parts) in the image, and thus it is able to not only estimate the number of the crowd, but also accurately locate the position of each individual to provide richer information about the spatial distribution rather than predicting an intermediate representation of the number as in the case of density maps. This strategy combines the advantages of target detection and point localisation and has received extensive attention and research from scholars in recent years.
As a result, these methods P2PNet ~\cite{song2021rethinking}, CLTR ~\cite{liang2022an}, PET ~\cite{liu2023pointquery}, APGCC ~\cite{chen2024improving} are known for their simplicity, end-to-end trainability, and lack of reliance on complex preprocessing and multi-scale feature map fusion. Although, point-based crowd counting methods have obvious advantages in terms of localisation accuracy and end-to-end training, we found that P2PNet's ~\cite{song2021rethinking} optimisation instability in matching point proposals to targets during training reduces the model's learning efficiency and counting accuracy; whereas, PET ~\cite{liu2023pointquery} can be limited by the size of the rectangular window, which results in leakage detection when dealing with large-sized targets; and APGCC ~\cite{chen2024improving} lacks the infrared light modality, which can greatly affects the performance of the model in bad scenes such as nighttime. Therefore, this paper aims to investigate a new method of bimodal feature fusion and point matching to improve the performance of crowd counting.

\subsection{Multi-spectral image fusion} 

Previously published studies aimed to address the question of where to fuse, i.e., which stage of fusion of input features to choose. Most of them explored the optimal fusion stage by designing macro-network architectures.Wagner et al ~\cite{wagner2016multispectral} investigated two deep fusion architectures (early fusion and late fusion) and analysed their performance on multispectral data. In order to exploit the complementary information of infrared and visible images, Liu et al ~\cite{liu2016multispectral} designed two other ConvNet fusion architectures (midway fusion and decision fusion) to improve the reliability of target detection and demonstrated that midway fusion enables the model to achieve the best detection performance. Based on this, ~\cite{fu2023lraf,shen2023icafusion,zhu2023multi,fang2021cross} introduced a Transformer-based fusion module in order to fuse more global complementary information between infrared and visible images. In addition to the direct fusion of image features, ~\cite{li2018illumination,zhou2020improving,yang2021baanet,li2024stabilizing} have used illumination-aware fusion methods to fuse IR and visible image features or post-fuse the results of multicrystalline system detection.~\cite{li2024stabilizing,li2023confidence} further use confidence or uncertainty scores of regions to post-fuse multibranch predictions. However, these methods neglect the modal misalignment problem, resulting in their inability to exploit misaligned object features. Therefore, this paper proposes a new adaptive fusion module to address the modal misalignment problem in infrared-visible crowd counting tasks.

\subsection{Multispectral Modal Misalignment Fusion} 

Modal misalignment is a critical problem in infrared visible object detection, and several recent works ~\cite{zhang2021weakly,yuan2024improving,yuan2022translation} have been devoted to solving this problem.Zhang et al ~\cite{zhang2021weakly} first solved the alignment problem by predicting the shift offsets of the reference proposal in another modality and fusing the alignment proposal features. ~\cite{yuan2024improving,yuan2022translation} further considered the scale and angle offsets of the reference proposal for more accurate alignment feature fusion in aerial target detection. ~\cite{yuan2024improving} calculates the attention value between feature points in the reference modality and another modality to achieve the fusion of unaligned object features. ~\cite{yuan2022translation} make full use of infrared and visible features to learn the intrinsic relationship between the same object in both modalities, and are able to output the exact position of the object in both modalities. However, these methods can only show good results on objects with larger targets and do not fully consider the problem of accurate alignment of the same object between the two modalities in dense scenes. In contrast, our method is capable of fine-grained fusion of infrared and visible images in dense scenes, fully multispectral features to learn the intrinsic relationship between the two modalities, and effectively align the potential spatial features between the modalities.


\section{A point-based crowd counting framework}

Previous work ~\cite{song2021rethinking} has demonstrated the effectiveness of an auxiliary point-guided crowd counting framework based on three main components: point proposal prediction, implicit feature interpolation, and auxiliary point-guided target matching.

\subsection{Point Proposal Projections}

The size of the depth feature map ${\mathcal{F}_s}$ output from the backbone network is $H \times W$, where $s$ denotes the downsampling step.The process consists of two main branches, regression and classification, which are used to predict the offset of the point coordinates and determine the confidence score, respectively. Specifically, ${\mathcal{F}_s}$ Each pixel on should correspond to a patch of input image size $s \times s$, in which a set of ${R_k} = \left( {{x_k},{y_k}} \right)$ with predefined positions is first introduced as fixed reference points $R = \left\{ {{R_k}|k \in \left\{ {1,...,K} \right\}} \right\}$, where $K$ represents the number of reference points. Thus the regression branch should generate ${R_k}$ point suggestions, assuming that the reference point ${\hat p_j} = \left( {{{\hat x}_j},{{\hat y}_j}} \right)$ predicts the offset $\left( {\Delta _{jx}^k,\Delta _{jy}^k} \right)$ of its point suggestion $\left( {\Delta _{jx}^k,\Delta _{jy}^k} \right)$, then the coordinates of ${\hat p_j}$ are computed as follows:
\begin{align}
	\begin{array}{l}
		{{\hat x}_j} = {x_k} + \gamma \Delta _{jx}^k,\vspace{10pt}\\
		{{\hat y}_j} = {y_k} + \gamma \Delta _{jy}^k
	\end{array}
\end{align}
where $\gamma $ is the offset of the scaling prediction.

\subsection{Implicit feature interpolation}
Since the auxiliary points are randomly assigned based on the ground truth coordinates, the traditional bilinear interpolation method is not suitable for extracting features at these arbitrary locations. Therefore, we propose to use implicit feature interpolation to obtain these features. Many studies ~\cite{park2019deepsdf,molaei2023implicit} have demonstrated that implicit functions show great potential in providing a continuous representation of features. This representation can capture more details and is beneficial for various computer vision tasks. In addition, implicit neural representations (INRs) approximate the signal function through a neural network, providing advantages over traditional representations, such as a representation that is no longer coupled to spatial resolution, high representational power, and high generalizability. Therefore, we utilize function-based implicit interpolation to extract arbitrary and robust potential feature representations.

For a given point coordinate $(x,y)$, the four closest potential features to it are denoted as $Z_i^*|i \in \left\{ {1,...,4} \right\}$. Their distances $\delta _i^*|i \in \left\{ {1,...,4} \right\}$ from the target potential feature are then computed. These four potential features and their computed distances are connected channel by channel, and this series of information is then fed into the MLP to produce the target potential feature. However, it is well known that MLPs tend to prioritize low-frequency information and usually ignore critical high-frequency details, which may affect the performance of MLPs ~\cite{basri2020frequency,rahaman2018spectral,tancik2020fourier}. To overcome this limitation, we adopt the location coding suggested in ~\cite{xu2021ultrasr}, which enhances the dimensionality of the distance information, thus addressing this loss of high-frequency details. The interpolated feature results for point $(x,y)$ are defined as follows:
\begin{align}
	{F_{proposal}}(x,y) = \sum\limits_{i = 1}^4 {\frac{{{S_i}}}{S}} {f_\theta }(Z_i^*,\delta _i^*,\phi (\delta _i^*)),
\end{align}
where ${S_i}$ denotes the area of the target point around the diagonal point, $S$ denotes the sum of the surrounding area, and is calculated as $S = \sum\limits_{i = 1}^4 {{S_i}} $, ${f_\theta }\left( . \right)$ denotes the MLP, and $\phi \left( . \right)$ denotes the location code.

\subsection{Auxiliary points to guide target matching}

\begin{figure*}[ht]
	\centering
	\includegraphics[width=\linewidth]{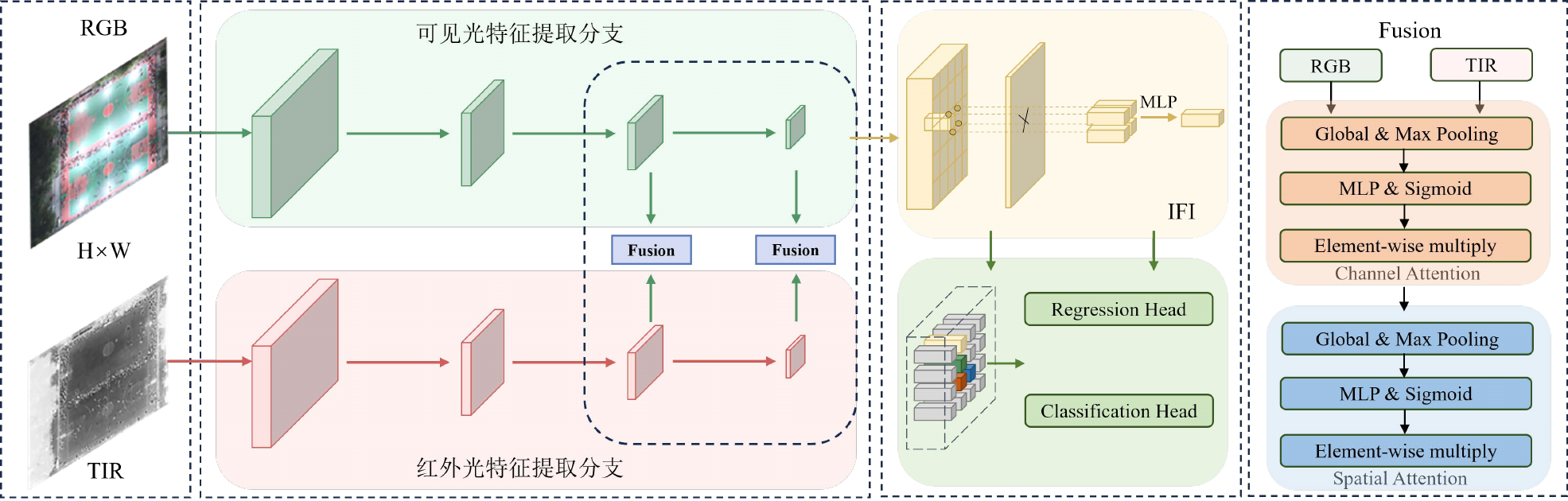}
	\caption{\textbf{Overall architecture of TAPNet}. We first extract the image feature representation $\left\{ {{F_{R1}},...,{F_{R4}}} \right\}$ and $\left\{ {{F_{T1}},...,{F_{T4}}} \right\}$ separately using the ResNet50 backbone. Then, a bi-optical attention fusion module is applied to the last two layers of features to fuse the features. Subsequently, the fused two layers of features $\left\{ {{F_3},{F_4}}\right\}$ are passed through an adaptive spatial pyramid pooling (ASPP) module and implicit feature interpolation (IFI), respectively. Finally, these features are cascaded and passed to a regression and classification module to obtain the coordinates and confidence of the final target head point, including “unoccupied” or “occupied” and its probability and localization.} 
	\label{fig:pipeline1}
	\vspace{-5pt}
\end{figure*}

In order to solve the instability problem in the target matching phase, we adopt the auxiliary point guidance mechanism suggested in ~\cite{chen2024improving}. As shown in Fig.~\ref{fig:pipeline1} the overall architecture of TAPNet, the set of auxiliary positive and negative points can be determined based on the point coordinates $(x,y)$, respectively:
\begin{align}
	\begin{array}{l}
		A_{{\rm{pos}}}^i = \{ (x + R_{{\rm{pos}}}^{i,x},y + R_{{\rm{pos}}}^{i,y})|i = 1,2,...,{k_{{\rm{pos}}}}\},\vspace{10pt}\\
		A_{{\rm{neg}}}^j = \{ (x + R_{{\rm{neg}}}^{j,x},y + R_{{\rm{neg}}}^{j,y})|j = 1,2,...,{k_{{\rm{neg}}}}\}, 
	\end{array}
\end{align}

Here, $R_{\text{pos}}^{i,x}$ and $R_{\text{pos}}^{i,y}$ represent a series of random numbers used to generate the $x$ and $y$ coordinates of positive points, each number uniformly distributed between $[-n_{\text{pos}}, n_{\text{pos}}]$. $k_{\text{pos}}$ and $k_{\text{neg}}$ represent the number of positive and negative points generated, respectively. Each set of $R_{\text{pos}}^{i}$ and $R_{\text{neg}}^{j}$ is used to create a unique set of coordinates for $A_{\text{pos}}$ and $A_{\text{neg}}$, and then these random numbers are used to offset the real position $(x,y)$. Features of auxiliary positive points are extracted to calculate the predicted positional confidence $\hat{c}_{\text{pos}}^{*}$ and the bias $\Delta_{\text{neg}}^{*}$, and then the position of each proposed point $\hat{p}_{\text{pos}}^{*}$ is computed.

To achieve one-to-one matching between predicted and real points, we use the Hungarian algorithm ~\cite{kuhn1955hungarian} as a proposal-target matching strategy,where $\Omega(P,\hat{P},D)$ assigns a real target from  $\hat{\mathcal{P}}$ to each point proposal in $\mathcal{P}$. To evaluate the distance between real and target points, a cost matrix of shape ${\hat p_j}$ is defined by combining the Euclidean distance between point-to-points and the confidence score ${\hat c_j}$ of each proposal $N \times M$:
\begin{align}
	\mathcal{D}(\mathcal{P},\widehat{\mathcal{P}})=\left(\tau\|p_i-\hat{p}_j\|_2-\hat{c}_j\right)_{i\in N,j\in M},
\end{align}

Among them, $\tau$ is a weighting factor, used to balance the effect of the pixel distance. $||\cdot||_2$ represents the $l_2$ distance. Note that, to ensure the predicted point number $M$ is greater than the number of true points $N$, enough matches can be generated. After the matching is completed, each true point $p_i$ is matched with a proposed point $\hat{p}_{\xi(i)}$ to obtain the optimal matching, denoted as $\xi = \Omega(\mathcal{P}, \hat{\mathcal{P}}, \mathcal{D})$. Therefore, the set of matched proposals is defined as $\hat{\mathcal{P}}_{pos} = \{\hat{p}_{\xi(i)} \mid i \in \{1, ..., N\}\}$ for positive matches, and the set of unmatched proposals is defined as $\hat{\mathcal{P}}_{neg} = \{\hat{p}_{\xi(i)} \mid i \in \{N + 1, ..., M\}\}$ for negative matches.

\section{Proposed Method}

In this work, to demonstrate the effectiveness of our proposed method, we extend the point-based APGCC ~\cite{chen2024improving} framework for multispectral crowd counting. Specifically, we propose a dual-optical attention fusion module (DAFP) that enhances modal fusion and interaction from both channel and spatial aspects by utilizing complementary information between multimodal images. In this paper, we also introduce an adaptive bi-optical feature decomposition fusion module (AFDF) to solve the problem of inaccurate localization caused by misalignment of feature fusion between two modalities.

\subsection{Architecture Overview}

The method architecture of this paper is shown in Fig.~\ref{fig:pipeline1} The overall architecture of TAPNet, which contains four main components: feature fusion module, BackBone, spatial random offset data enhancement, and auxiliary points to guide target matching.

Many studies ~\cite{yuan2023c2former,liang2022an,guo2024dpdetr} have demonstrated that ResNet50 performs well as a backbone network in a variety of deep learning tasks. Therefore, in this paper, we use ResNet50 as a backbone network for extracting fused image feature representations.As shown in Figure~\ref{fig:pipeline1}, for the input infrared image $\mathcal{I}\in \mathbb{R}^{H\times W\times 3}$ and visible image $V\in \mathbb{R}^{H\times W\times 3}$, ResNet50 is used to extract four levels of visible features $\{F_{r1}, F_{r2}, F_{r3}, F_{r4}\}$ and infrared features $\{F_{t1}, F_{t2}, F_{t3}, F_{t4}\}$. The last two layers of features are fused through the Dual Attention Fusion (DAF) module to obtain the fused features $\{F_3, F_4\}$. In contrast, the Adaptive Feature Decomposition Fusion (AFD) module adopts early fusion, meaning that the fused image can be processed through a single Backbone to obtain the last two layers of feature maps. Subsequently, the fused two layers of features $\{F_3, F_4\}$ are passed through an Adaptive Spatial Pyramid Pooling (ASPP) module and an Implicit Feature Interpolation (IFI) module respectively, to compute the feature $F_{proposal}(x, y)$. This allows the model to smoothly transition between different scales for more coherent feature representation. Finally, these features are concatenated and passed into regression and classification modules to obtain the coordinates and confidence scores of the target head points.

\begin{figure*}[ht]
	\centering
	\includegraphics[width=\linewidth]{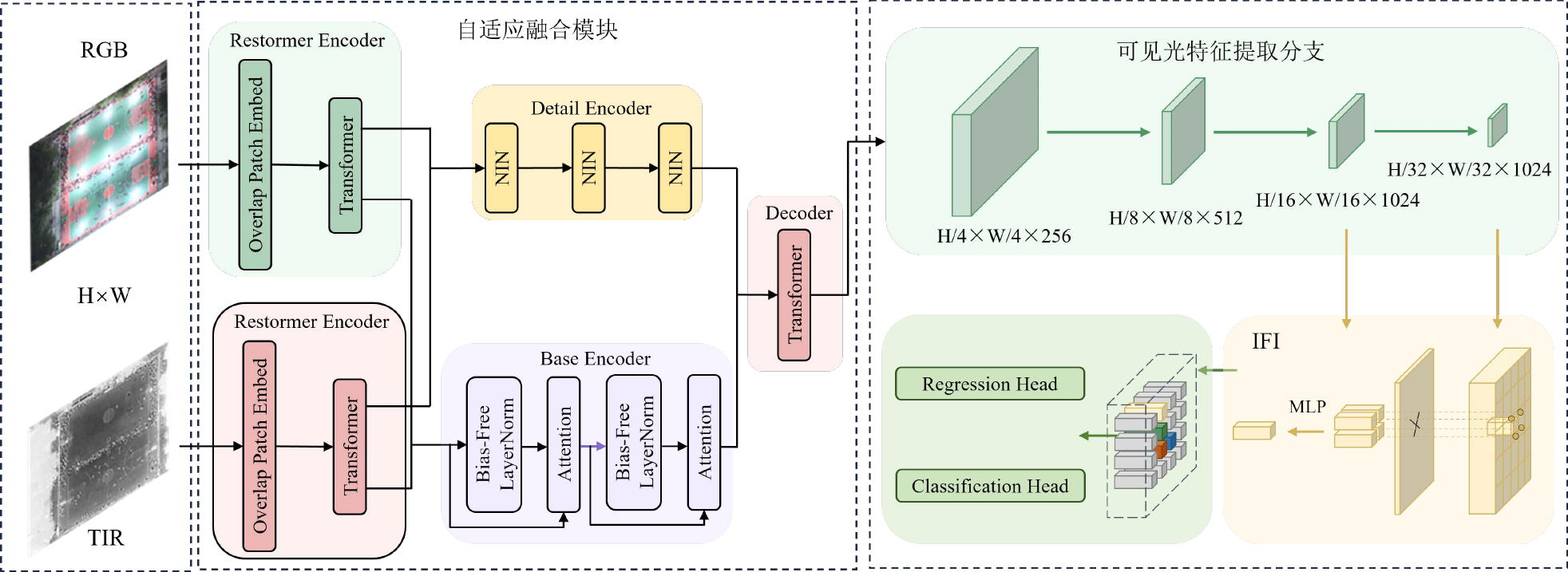}
	\caption{\textbf{Adaptive Fusion Architecture Diagram}. The module consists of an encoder and decoder, respectively, and a domain-adaptive layer structure based on hybrid kernel functions. The difference with Figure~\ref{fig:pipeline1} lies in the fact that the Adaptive Fusion Module for Bi-Optical Feature Decomposition (AFD) employs early fusion, which also means that the fused image is passed through BackBone once to obtain the last two layers of the feature map.} 
	\label{fig:pipeline2}
	\vspace{-5pt}
\end{figure*}

\subsection{Attention Fusion Module}
\label{sec:rect_win}  

In order to realize the effective fusion of the two modalities, it is inspired by the feature-enhanced long-range attention fusion network Fig.~\ref{fig:pipeline1} based on feature enhancement proposed by Fu et al ~\cite{fu2023lraf}. In this paper, we design a feature fusion module (DAF) based on the dual-attention mechanism, as shown in Fig. 2.1, which enhances modal fusion and interaction from both channel and spatial aspects by utilizing complementary information between multimodal images.

\vspace{-5pt}
\paragraph{Channel Attention Branching.} 
For the given inputs $F_R \in \mathbb{R}^{C \times H \times W}$ and $F_T \in \mathbb{R}^{C \times H \times W}$, global polling (Global Polling) and max pooling operations are first performed through channel attention. Therefore, the output of the pooling operation can be represented as follows:
\begin{align}
		F_{R_{\text{avg}}}(c) = \frac{1}{H \times W} \sum_{i=1}^{H} \sum_{j=1}^{W} F_R(c, i, j),
\end{align}
\begin{align}
	F_{R_{\text{max}}}(c) = \max_{i, j} F_R(c, i, j), 
\end{align}
Subsequently, the pooling results of RGB and TIR are concatenated to form a joint representation to capture the complementary information between modalities, which can be expressed as:
\begin{align}
	F_{\text{cat}} = [F_{R_{\text{avg}}}, F_{R_{\text{max}}}, F_{T_{\text{avg}}}, F_{T_{\text{max}}}]
\end{align}
Then, the concatenated features are passed through a shared multi-layer perceptron (MLP), and the Sigmoid activation function is used to generate weights for each channel:
\begin{align}
	\begin{array}{l}
		w_{c1} = \sigma(\text{MLP}(F_1(F_{\text{cat}}))),\vspace{10pt} \\
		w_{c2} = \sigma(\text{MLP}(F_2(F_{\text{cat}}))), 
	\end{array}
\end{align}
Here, $F_1(\cdot)$ and $F_2(\cdot)$ are respectively a 1x1 convolutional layer, and $\sigma$ is the Sigmoid activation function. Finally, the generated weights are applied to the inputs to obtain the final fused feature map, which can be represented as:
\begin{equation}
	F_R^{\prime} = w_{c1} F_R, \\
	F_T^{\prime} = w_{c2} F_T,
\end{equation}

\vspace{-10pt}
\paragraph{Branching of spatial attention.}
In order to fully exploit the complementarity between visible and infrared modalities, after the channel attention branches, we further enhance the complementarily enhanced features through spatial fusion. Similar to the channel attention branches, this branch also first performs global polling (Global Polling) and max pooling (Max Pooling) operations, with the output represented as:
\begin{align}
	\begin{array}{l}
		F'_{R_{\text{avg}}}(i,j) = \frac{1}{C} \sum_{c=1}^{C} F_R(c,i,j),\vspace{10pt} \\
		F'_{R_{\text{max}}}(i,j) = \max_{c} F_R(c,i,j), 
	\end{array}
\end{align}
Subsequently, the pooling results of the visible and infrared modalities are concatenated along the channel dimension to obtain:
\begin{equation}
	F'_{\text{cat}} = [F'_{R_{\text{avg}}}, F'_{R_{\text{max}}}, F'_{T_{\text{avg}}}, F'_{T_{\text{max}}}]
\end{equation}
Then, a 1x1 convolutional layer and a Sigmoid activation function are used to generate spatial attention weights:
\begin{align}
	\begin{array}{l}
		w'_{c2} = \sigma(F'_4(F'_{\text{cat}})),\vspace{10pt} \\
		w'_{c2} = \sigma(F'_4(F'_{\text{cat}})), 
	\end{array}
\end{align}
Here, $F'_3(\cdot)$ and $F'_4(\cdot)$ are respectively 1x1 convolutional layers. Finally, the generated weights are multiplied with the input features to obtain:
\begin{equation}
	F = \alpha \cdot w'_{c1} F_R + \beta \cdot w'_{c2} F_T
\end{equation}
where $\alpha$ and $\beta$ are the modality fusion weights, usually initialized to equal values, but can also be adjusted dynamically through learning.

\subsection{Adaptive Fusion Module}

In order to better align the latent spatial features of infrared and visible images, inspired by the work of ~\cite{xu2024daf}, an adaptive two-branch feature decomposition fusion module is proposed in this paper. As shown in Figure~\ref{fig:pipeline2}, the module consists of an encoder and decoder, respectively, as well as a domain adaptive layer structure based on hybrid kernel functions.

\vspace{-10pt}
\paragraph{Twin-branch encoder module.}
The dual-branch encoder module is designed to simultaneously process global structural information and detailed texture information for efficient fusion between RGB images and infrared images (TIR). The encoder mainly consists of a Transorfmer shared layer, a base encoder and a detail encoder, and by utilizing the Transformer's self-attention mechanism, the network can perform both intra- and inter-modal fusion and robustly capture potential interactions between RGB and TIR. The base encoder is based on the Restormer network, which is responsible for capturing global structural information, while the detail encoder is based on an invertible neural network (INN), which is responsible for extracting detailed texture information.

For the given input visible and infrared images (denoted as \(I \in \mathbb{R}^{H \times W \times 3}\) and \(V \in \mathbb{R}^{H \times W \times 3}\)), after feature extraction through the Transformer shared layer, we obtain:
\begin{equation}
	Y_I^S = E_S(I), Y_V^S = E_S(V),
\end{equation}
where \(E_S(\cdot)\) represents the Transformer shared layer, which consists of multiple Transformer modules, each containing a self-attention mechanism and a feedforward network. To enhance the interaction between features, we introduce depthwise separable convolutions and a temperature parameter \(\tau\) to dynamically adjust the scaling of attention scores:
\begin{equation}
	Attention(Q, K, V) = softmax\left(\frac{QK^T}{\tau}\right)V
\end{equation}
here, \(Q\), \(K\), \(V\) are query, key, and value matrices generated through convolution operations. This design retains the global modeling capability of Transformers and enhances the perception of spatial information in images through convolution operations.

We input the obtained features into two encoders to get:
\begin{align}
	\begin{array}{ll}
		Y_I^B = E_B(Y_I^S), \quad Y_V^B = E_B(Y_V^S),\vspace{10pt} \\
		Y_I^D = E_D(Y_I^S), \quad Y_V^D = E_D(Y_V^S),
	\end{array}
\end{align}
where \(E_B(\cdot)\) and \(E_D(\cdot)\) represent the base encoder and detail encoder, respectively. Next, after cross fusion of the obtained features, they are input into the two encoders again, expressed as:
\begin{align}
	\begin{array}{ll}
		Y_{MV}^B &= F_B(Y_I^B + Y_V^B),\vspace{10pt} \\
Y_{MV}^D &= F_D(Y_I^D + Y_V^D)
	\end{array}
\end{align}
Here, \(F_B(\cdot)\) and \(F_D(\cdot)\) are the base encoder and detail encoder. Additionally, to ensure global feature consistency, avoid overalignment of local details, and prevent loss of modality-specific information, MK-MMD is only applied in the base encoder, enabling the model to better balance global structure and detail retention.

\vspace{-10pt}
\paragraph{Decoder Module.}

The decoder part is then used to better fuse the results of the two-branch encoder and to provide image support for the following point matching network. Therefore, the missing IR features on the visible light need to be fused with the visible light and reshaped to get a new visible fusion image:

\begin{equation}
	F_A = D\left(V, Y_{MV}^B, Y_{MV}^D\right)
\end{equation}
where $D(\cdot)$ denotes the decoder module, which employs Transformer blocks as fundamental components and leverages residual connections to integrate the visible image V into the output reshaped image.

\vspace{-10pt}
\paragraph{Adaptive layer.}

In order to solve the problem of distribution difference between infrared and visible images, we introduce Multi-Kernel Maximum Mean Difference (MK-MMD) to self-adaptively regulate the fusion between the two modalities. The core idea is to achieve feature alignment of the bimodal fused images by mapping the images into a kernel Hilbert space (RKHS) and minimizing the distributional differences in the shared feature space using a hybrid kernel function.
MK-MMD was developed based on the original MMD and proposed by Gretton in 2012. One of the most important concepts is the kernel function, which is fixed in the traditional MMD and mostly uses Gaussian kernel, denoted as:
\begin{equation}
	 k_G(x_i^t, x_i^v) = \sum_{j=1}^{K} \alpha_j \exp \left( -\frac{||x_i^t - x_i^v||^2}{2\tau_j^2} \right),
\end{equation}
where \(x_i^t\) and \(x_i^v\) represent the \(i\)-th sample of the infrared and visible images, respectively. \(\tau_j\) is the bandwidth of the \(j\)-th Gaussian kernel, controlled by the hyperparameter \(\gamma\) as \(\tau = 1/\sqrt{2\gamma}\). \(\alpha_j\) is the weight of the \(j\)-th kernel (usually non-negative and summing up to 1), \(K\) denotes the number of Gaussian kernels. Unlike the Gaussian kernel, Laplacian kernel is more sensitive to edges and defined as:
\begin{equation} 
	k_L(x_i^t, x_i^v) = \sum_{j=1}^{K} \beta_j \exp \left( -\frac{||x_i^t - x_i^v||}{\tau_j} \right),
\end{equation}
Here, \(\beta_j\) is the weight of the \(j\)-th kernel (also non-negative and summing up to 1). To address the issue of selecting a single kernel, MK-MMD proposes constructing a unified kernel using multiple kernels. This paper combines the Gaussian and Laplacian kernels to propose a hybrid multi-kernel maximum mean discrepancy, i.e.,
\begin{equation}
	k_H(x_i^t, x_i^v) = c_1 k_G(x_i^t, x_i^v) + c_2 k_L(x_i^t, x_i^v),
\end{equation}
where \(c_1\) and \(c_2\) denote the weights of the Gaussian and Laplacian kernels respectively, both are set to 1.0 in this paper. The hybrid multi-kernel is capable of capturing both global structures and local details differences between infrared and visible images more comprehensively.

To project the infrared feature \( F_t \) and visible feature \( F_v \) into the reproducing kernel Hilbert space and minimize the distribution discrepancy in the shared feature space, it is expressed as:
\begin{equation}
	d_{kl}(S_t, S_v) = \left\| \mathbb{E}_{x_t}[F_t] - \mathbb{E}_{x_v}[F_v] \right\|_{\mathcal{H}_k}^2,
\end{equation}
where \(\mathbb{E}[\cdot]\) denotes the expectation. By incorporating the hybrid multi-kernel maximum mean discrepancy, the quality of fused images under complex scenarios is enhanced.

\subsection{Spatial Random Offset Data Enhancement}

To enhance the model's counting ability on misaligned dual-source images, we propose a spatial random shift data augmentation strategy. Specifically, the training data is feature-aligned using a Generative Adversarial Network (GAN), resulting in a different sampling distribution compared to the unaligned validation set. To address this discrepancy, we apply random horizontal and vertical shifts ($\Delta x$ and $\Delta y$) to the TIR image $V(x, y)$, expressed as:
\begin{equation}
	V'(x, y) = V(x - \Delta x, y - \Delta y)
\end{equation}
where $\Delta x$ and $\Delta y$ are sampled from $\Delta x, \Delta y \leftarrow \text{Rand}(-10, 10)$. This strategy improves the model's accuracy with misaligned dual-source images, especially in crowd counting tasks, by simulating random object movements in the image.


\begin{table*}[htbp]
	\centering
	\caption{People Counting Dataset}
	\begin{tabular}{l l l l l l l l l}
		\specialrule{1pt}{0pt}{0pt} 
		\textbf{Dataset} & \textbf{Type} & \textbf{Resolution} & \multicolumn{3}{c}{\textbf{Image Num}} & \multicolumn{3}{c}{\textbf{People Num}} \\
		& & & \textbf{Train} & \textbf{Val} & \textbf{Sum} & \textbf{Min} & \textbf{Max} & \textbf{Total} \\ \hline
		DroneRGBT & RGB-TIR & 512×640 & 1807 & 1800 & 3607 & 1 & 403 & 175698 \\
		GAIIC2 & RGB-TIR & 512×640 & 1807 & 1000 & 2807 & 1 & 407 & 29780 \\
		\specialrule{1pt}{0pt}{0pt} 
	\end{tabular}
	\label{table:dataset}
\end{table*}

\subsection{Calculation of losses}

In our training process, the loss function is divided into: adaptive fusion loss, auxiliary point loss, and regression classification loss composition.

\vspace{-10pt}
\paragraph{Adaptive fusion loss.}

In the encoder-decoder stage, the loss function \( \mathcal{L}_{ed} \) is composed of correlation loss, decomposition loss, MK-MMD loss, and InfoNCE loss. To capture the cross-modal relationships, we introduce the correlation loss \( \mathcal{L}_{cc} \), which balances the correlation between global features and detailed features, as shown below:
\begin{align}
	\begin{array}{ll}
		\mathcal{L}_{cc}^B = C(Y_I^B, Y_I^B),\vspace{10pt} \\
		\mathcal{L}_{cc}^D = C(Y_I^D, Y_I^D),
	\end{array}
\end{align}
where \( C(\cdot) \) represents the correlation operation. Additionally, we introduce a Decomposition Loss to further refine the correlation of fine-grained features. The calculation formula for decomposition loss is as follows:
\begin{equation}
	\mathcal{L}_{cc} = \alpha \frac{\mathcal{L}_{cc}^{D^2}}{1.01 + \mathcal{L}_{cc}^B},
\end{equation}

To align the feature distributions of different modalities, we employ the constructed hybrid multi-kernel to calculate the distribution discrepancy between low-frequency features of infrared and visible images, and compute the MK-MMD loss, expressed as:
\begin{equation}
	\mathcal{L}_{\text{mmd}} = d_{kl}(Y_I^B, Y_V^B),
\end{equation}

The InfoNCE loss is also utilized in the training process. It helps the model learn semantically meaningful features by contrasting positive sample pairs (from the same class) and negative sample pairs (from different classes). It is defined as:
\begin{equation}
	\mathcal{L}_{\text{InfoNCE}} = -\frac{1}{K} \sum_{i=1}^{K} \log \frac{\exp\left(\frac{\text{sim}(x_i, y_i)}{\tau}\right)}{\sum_{j=1}^{K} \exp\left(\frac{\text{sim}(x_i, y_j)}{\tau}\right)},
\end{equation}
where \(K\) is the batch size, \(\text{sim}(x_i, y_j)\) measures the similarity between feature vectors \(x_i\) and \(y_j\) (here, the dot product is used), and \(\tau\) is a temperature parameter set to 0.1 in this context. The loss function encourages positive pairs to have similar feature vectors while pushing negative pairs apart, thus learning better feature representations.

Consequently, the loss function during the encoder-decoder training phase is expressed as:
\begin{equation}
	\mathcal{L}_{\text{ed}} = \beta_1 \mathcal{L}_{cc}^B + \beta_2 \mathcal{L}_{cc}^D + \beta_3 \mathcal{L}_{\text{mmd}} + \beta_4 \mathcal{L}_{\text{InfoNCE}},
\end{equation}
where \(\beta_1\), \(\beta_2\), \(\beta_3\), and \(\beta_4\) denote the respective weighting parameters.

During the fusion layer training phase, the loss function \( \mathcal{L}_{\text{fusion}} \) is composed of intensity loss, maximum gradient loss, and decomposition loss. Intensity loss is commonly used to measure the consistency in pixel intensity between the fused image and the reference image. The typical intensity loss is the Mean Squared Error (MSE) loss, formulated as:
\begin{equation}
	\mathcal{L}_{\text{in}} = \frac{1}{L} \sum_{i=1}^{L} \left\| \max(Y_i, I_i) - \hat{F}_i \right\|_1
\end{equation}
where \( L \) denotes the total number of pixels, and \( \left\| \cdot \right\|_1 \) represents the \( L_1 \) norm, i.e., the sum of absolute values. Maximum gradient loss is utilized to preserve edge information, ensuring that the edges of the fused image align with those of the reference image, expressed as:
\begin{equation}
	\mathcal{L}_{\text{max\_grad}} = \frac{1}{L} \sum_{i=1}^{L} \left\| \max(\nabla V_i, \nabla I_i) - \nabla \hat{F}_i \right\|_1
\end{equation}
here, \( \nabla V_i \) and \( \nabla I_i \) respectively represent the gradient values of the two input images at the \( i \)-th pixel point, while \( \nabla \hat{F}_i \) denotes the gradient value of the fused image at the \( i \)-th pixel point.

Thus, the loss function for the fusion phase is obtained as:
\begin{equation}
	\mathcal{L}_{\text{fuse}} = \mathcal{L}_{\text{in}} + \gamma_1 \mathcal{L}_{\text{max\_grad}} + \gamma_2 \mathcal{L}_{\text{cc}}
\end{equation}
where \( \gamma_1 \) and \( \gamma_2 \) respectively denote the weighting parameters. The overall adaptive fusion loss is:
\begin{equation}
	\mathcal{L}_{\text{af}} = \mathcal{L}_{\text{ed}} + \mathcal{L}_{\text{fuse}}
\end{equation}

\vspace{-10pt}
\paragraph{Loss of auxiliary points.}

In addition, it is necessary to determine the loss of the auxiliary points. Our goal is to ensure that the confidence of the auxiliary positive points is as close to 1 as possible and that their predicted displacement is as close to zero as possible in terms of Euclidean distance. To achieve this, we define the loss function for the auxiliary positive points as follows:
\begin{equation}
	\begin{split}
		\mathcal{L}_{pos} &= 
		\frac{1}{N} \frac{1}{k_{pos}}
		\sum_{l=1}^{N} 
		\sum_{i=1}^{k_{pos}} 
		\Big( \log c_{pos}^{*}(l, i) \\ 
		&+ \lambda_1 \| p_i - \hat{p}_{pos}^{*}(l, i) \|_2^2
		\Big),
	\end{split}
\end{equation}
where \(\lambda_1\) denotes the proportionality factor. For auxiliary negative points, our aim is to ensure that their confidence \(\hat{c}_{neg}^{*}\) and displacement \(\Delta_{neg}^{*}\) are as close to zero as possible. This prevents negative points from using displacement to bring their proposal coordinates close to true values, which is crucial for reducing the likelihood of these negative points being incorrectly regarded as matched proposals during the matching process. The loss function for the auxiliary negative points is defined as:
\begin{equation}
	\begin{split}
		\mathcal{L}_{neg} &= 
		\frac{1}{N} \frac{1}{k_{neg}}
		\sum_{l=1}^{N} 
		\sum_{j=1}^{k_{neg}} 
		\Big( \log (1 - \hat{c}_{neg}^{*}(l, j)) \\
		&+ \lambda_2 \| \Delta_{neg}^{*}(l, j) \|_2^2 
		\Big),
	\end{split}
\end{equation}
where \(\lambda_2\) represents the proportionality factor. Consequently, the total loss guided by the auxiliary points can be expressed as:
\begin{equation}
	\mathcal{L}_{apg} = \mathcal{L}_{pos} + \mathcal{L}_{neg}
\end{equation}
Through this additional guidance, we can instruct the network to train the nearest point proposals as positive points while treating distant points as negative points. It is crucial that the selected positive points are likely correct matches and are very close to the true points.

\vspace{-10pt}
\paragraph{Regression classification loss.}

After obtaining the true target, we calculate the Euclidean loss \(\mathcal{L}_{\text{loc}}\) for point regression, the calculation formula is as follows:

\begin{equation}
	\mathcal{L}_{\text{loc}} = \frac{1}{N} \sum_{i=1}^{N} \| p_i - \hat{p}_{\text{gt}}(i) \|_2^2,
\end{equation}

We also use cross-entropy loss \(\mathcal{L}_{\text{pos}}\) for training proposal classification, the calculation formula is as follows:

\begin{equation}
	\mathcal{L}_{\text{ciz}} = -\frac{1}{M} \left\{ \sum_{i=1}^{N} \log \hat{c}_{\text{g}}(i) + \lambda_3 \sum_{i=N+1}^{M} \log \left( 1 - \hat{c}_{\text{s}}(i) \right) \right\},
\end{equation}

where \(\lambda_3\) represents the newly added weighting parameter for the negative proposals. Therefore, the total loss for regression and classification is defined as:

\begin{equation}
	\mathcal{L}_{\text{point}} = \mathcal{L}_{\text{ciz}} + \lambda_4 \mathcal{L}_{\text{loc}},
\end{equation}

where \(\lambda_4\) represents the weighting parameter for balancing regression loss. Overall, the total loss function for TAPNet is the sum of the three loss functions as follows:

\begin{equation}
	\mathcal{L} = \mathcal{L}_{\text{af}} + \mathcal{L}_{\text{apg}} + \mathcal{L}_{\text{point}},
\end{equation}


\begin{table*}[htbp]
	\centering
	\caption{Performance Comparison of Crowd Counting Models on DroneRGBT and GAII C2 Datasets Using RGB, TIR, and RGB-TIR Images for Training}
	\begin{tabular}{l l l l l l l l l l}
		\specialrule{1pt}{0pt}{0pt} 
		\textbf{Method} & \textbf{Backbone} & \textbf{Approach} & \textbf{Modality} & \multicolumn{3}{c}{\textbf{DroneRGBT}} & \multicolumn{3}{c}{\textbf{GAII C2}} \\
		& & & & \textbf{MAE$\downarrow$} & \textbf{MSE$\downarrow$} & \textbf{F1$\uparrow$} & \textbf{MAE$\downarrow$} & \textbf{MSE$\downarrow$} & \textbf{F1$\uparrow$} \\ 
		\hline
		P2PNet & Vgg16 & Point & RGB & 10.83 & 17.09 & 0.596 & 10.95 & 21.01 & 0.455 \\
		CLTR & ResNet50 & Point & RGB & 12.06 & 20.86 & 0.587 & 11.37 & 21.88 & 0.423 \\
		PET & Vgg16 & Point & RGB & 10.92 & 16.85 & \underline{0.611} & 10.10 & 17.36 & 0.412 \\
		APGCC & Vgg16 & Point & RGB & 11.50 & 16.61 & 0.603 & 10.35 & 18.92 & 0.409 \\
		\hline
		&  &  & RGB & 11.3 & 22.1 & - & 10.73 & 20.60 & 0.468 \\
		DroneNet & YOLOv5 & Detection & TIR & 18.6 & 25.2 & - & 15.86 & 25.62 & 0.379 \\
		&  &  & R-T & 10.1 & 18.8 & - & 9.93 & 17.39 & 0.491 \\
		\hline
		&  &  & RGB & 10.8 & 21.1 & - & 10.11 & 21.01 & 0.397 \\
		MMCount & CNN & Map & TIR & 16.0 & 23.3 & - & 15.25 & 22.82 & 0.334 \\
		&  &  & R-T & \underline{9.2} & 18.0 & - & 9.78 & 19.33 & 0.489 \\
		\hline
		&  &  & RGB & 10.32 & \underline{16.14} & 0.610 & \underline{8.54} & \underline{13.63} & \underline{0.506} \\
		TAPNet (ours) & ResNet50 & Point & TIR & 13.15 & 19.86 & 0.586 & 13.91 & 20.06 & 0.465 \\
		&  &  & R-T & \textbf{7.32} & \textbf{11.54} & \textbf{0.657} & \textbf{7.87} & \textbf{13.25} & \textbf{0.526} \\
		\specialrule{1pt}{0pt}{0pt} 
	\end{tabular}
	\label{table:sota}
\end{table*}


\section{Datasets and Implementation Details}

\subsection{Datasets} 
\label{sec:implementation}

In this paper, the effectiveness of the proposed method is evaluated on two challenging public datasets, DroneRGBT ~\cite{peng2020rgb} and GAIIC2, and all details of the datasets are detailed in Table~\ref{table:dataset}

The DroneRGBT dataset is a UAV-based RGB-Thermal crowd counting dataset first proposed by the Machine Learning and Data Mining Laboratory of Tianjin University ~\cite{peng2020rgb} in 2020, which contains 3607 pairs of RGB and TIR images, all of which have a fixed resolution (512 × 640), and 1807 pairs are respectively used for training, and 1800 pairs correspond to the Validation.
The GAIIC2 dataset, which is provided by the 2024 Global Artificial Intelligence Technological Innovation Competition, contains 2807 pairs of RGB (red, green and blue) and TIR (thermal infrared) images, with 1807 pairs assigned for training and 1000 pairs for validation. In order to evaluate the performance of the algorithms in real applications, this paper manually annotates the RGB and TIR images of the 1000-pair validation set in GAII24, respectively.

\subsection{Evaluation indicators} 

Following the conventions of existing work on population counting ~\cite{chen2024effectiveness,tota2015counting,zhang2016single}, this paper uses Mean Absolute Error (MAE), Mean Squared Error (MSE), and F1-Score for evaluating the counting performance of the model in this paper. Specifically, the Mean Absolute Error (MAE) is the average of the absolute difference between the predicted counts and the true counts, which provides an intuitive measure of how much the predicted values deviate from the true values. The mean square error (MSE) is the average of the squared prediction errors, which gives higher weight to larger errors and requires the model to be robust to noise and outliers in the data.MAE is defined as the average absolute error between predicted and actual values:
\begin{equation}
	MAE = \frac{1}{N} \sum_{i=1}^{N} \left| c_i - \hat{c}_i \right|
\end{equation}
MSE, representing the mean squared error, assesses the overall stability of the algorithm on the dataset:
\begin{equation}
	MSE = \sqrt{\frac{1}{N} \sum_{i=1}^{N} \left( c_i - \hat{c}_i \right)^2}
\end{equation}
Here $N$ denotes the number of test images, $c_i$ and $\hat{c}_i$ are the predicted and ground truth people counts for the $i$-th image, respectively. In summary, MAE indicates the accuracy and generalization ability of the counting algorithm, while MSE signifies the robustness of the algorithm across the dataset.

To further evaluate the accuracy of model predictions for crowd localization, we introduce the F1-Score as a comprehensive metric for assessing crowd counting and positioning models, defined as follows:
\begin{equation}
	F_1 = \frac{2 \times AP \times AR}{AP + AR}
\end{equation}
Where AP (Average Precision) and AR (Average Recall) are defined as:
\begin{equation}
	AP = \frac{\sum_{k=1}^{K} P(k) \cdot \delta(k)}{\sum_{k=1}^{K} \delta(k)}, \quad AR = \frac{\sum_{k=1}^{K} R(k) \cdot \delta(k)}{\sum_{k=1}^{K} \delta(k)}
\end{equation}
In these, $P(k)$ and $R(k)$ are the precision and recall at threshold $k$, respectively, and $\delta(k)$ is an indicator function that equals 1 if at least one true instance is detected at threshold $k$, otherwise 0, where $K$ is the total number of thresholds. Here, the matching between predicted and ground truth points uses the Hungarian matching algorithm for one-to-one matching.


\begin{table*}[htbp]
	\centering
	\caption{Ablation Results of the Dual Fusion Module}
	\begin{tabular}{lcccccc}
		\specialrule{1pt}{0pt}{0pt} 
		\textbf{Method} & \multicolumn{3}{c}{\textbf{DroneRGBT}} & \multicolumn{3}{c}{\textbf{GAII C2}} \\
		& \textbf{MAE$\downarrow$} & \textbf{MSE$\downarrow$} & \textbf{F1$\uparrow$} & \textbf{MAE$\downarrow$} & \textbf{MSE$\downarrow$} & \textbf{F1$\uparrow$} \\ 
		\hline
		RGB & 10.32 & 16.14 & 0.610 & 8.54 & 13.63 & 0.506 \\
		TIR & 13.15 & 19.86 & 0.586 & 13.91 & 20.06 & 0.465 \\
		R-T & 7.96 & 13.78 & 0.659 & 9.32 & 14.65 & 0.437 \\
		R-T+DAFP & \textbf{7.32} & \textbf{11.71} & 0.697 & 8.03 & 13.92 & 0.506 \\
		R-T+AFDF & 7.51 & 12.06 & \textbf{0.712} & \textbf{7.92} & \textbf{13.38} & \textbf{0.523} \\
		\specialrule{1pt}{0pt}{0pt} 
	\end{tabular}
	\label{table:dual_fusion}
\end{table*}


\begin{table}[htbp]
	\centering
	\caption{Evaluation of Head Points vs. Box Counts}
	\begin{tabular}{lccc}
		\specialrule{1pt}{0pt}{0pt} 
		\textbf{Method} & \textbf{MAE$\downarrow$} & \textbf{MSE$\downarrow$} & \textbf{F1$\uparrow$} \\ 
		\hline
		Boxes & 8.98 & 14.12 & 0.625 \\
		Point & \textbf{7.32} & \textbf{11.54} & \textbf{0.657} \\
		\specialrule{1pt}{0pt}{0pt} 
	\end{tabular}
	\label{table:Point}
\end{table}


\begin{table}[htbp]
	\centering
	\caption{Evaluation of Different Auxiliary Point Quantities}
	\begin{tabular}{lcccccc}
		\specialrule{1pt}{0pt}{0pt} 
		\textbf{(k$_{p}$, k$_{n}$)} & (0, 0) & (1, 0) & (1, 1) & (2, 0) & (2, 2) & (5, 5) \\
		\hline
		\textbf{MAE$\downarrow$} & 7.53 & \textbf{7.32} & 7.54 & 7.83 & 7.69 & 7.65 \\
		\textbf{MSE$\downarrow$} & 11.86 & \textbf{11.54} & 12.19 & 12.24 & 12.12 & 11.95 \\
		\textbf{F1$\uparrow$} & 0.635 & \textbf{0.657} & 0.628 & 0.625 & 0.631 & 0.625 \\
		\specialrule{1pt}{0pt}{0pt} 
	\end{tabular}
	\label{table:point_nums}
\end{table}


\begin{table}[htbp]
	\centering
	\caption{Evaluation of Different Auxiliary Point Random Ranges}
	\begin{tabular}{lccccc}
		\specialrule{1pt}{0pt}{0pt} 
		\textbf{(n$_{pos}$, n$_{neg}$)} & (1, 4) & (2, 8) & (3, 12) & (4, 16) \\
		\hline
		\textbf{MAE$\downarrow$} & 7.32 & 7.37 & 7.50 & 7.65 \\
		\textbf{MSE$\downarrow$} & 11.54 & 11.73 & 11.65 & 12.62 \\
		\textbf{F1$\uparrow$} & 0.657 & 0.628 & 0.6335 & 0.641 \\
		\specialrule{1pt}{0pt}{0pt} 
	\end{tabular}
	\label{table:point_random}
\end{table}


\begin{table}[htbp]
	\centering
	\caption{Comparison of Model Complexity and Performance}
	\begin{tabular}{lcc}
		\specialrule{1pt}{0pt}{0pt} 
		\textbf{Method} & \textbf{DAFP(ours)} & \textbf{AFDF(ours)} \\
		\hline
		\textbf{Parameters (M)$\downarrow$} & 32.75 & 32.98 \\
		\textbf{Inference Time (s)$\downarrow$} & 0.0296 & 0.059 \\
		\specialrule{1pt}{0pt}{0pt} 
	\end{tabular}
	\label{table:model_eva}
\end{table}


\begin{table}[htbp]
	\centering
	\caption{Ablation Results of Spatial Random Shift Data Augmentation Strategy}
	\begin{tabular}{lccc}
		\specialrule{1pt}{0pt}{0pt} 
		\textbf{Method} & \textbf{MAE$\downarrow$} & \textbf{MSE$\downarrow$} & \textbf{F1$\uparrow$} \\
		\hline
		DAFP & 9.23 & 14.32 & 0.465 \\
		DAFP+Spatial Shift & 7.96 & 13.55 & 0.512 \\
		AFDF & 8.98 & 14.16 & 0.491 \\
		AFDF+Spatial Shift & \textbf{7.87} & \textbf{13.25} & \textbf{0.526} \\
		\specialrule{1pt}{0pt}{0pt} 
	\end{tabular}
	\label{table:spatial_offset}
\end{table}
\subsection{Training details} 
\label{sec:experiment}

For data enhancement, we use the large-scale jitter (LSJ) enhancement method ~\cite{du2021simple,ghiasi2020simple} with random scaling (scaling factor range: [0.7, 1.3], ensuring that the shorter side is at least 128 pixels, and then the scaled image is randomly cropped into four fixed 128 × 128 pixel blocks and randomly flipped using a probability of 0.5. For the offset GAIIC2 dataset, we use the spatially randomized offset data enhancement strategy proposed in 2.4, which enables the validation set to have the same image offset distribution as the training set. Data augmentation is only used in the two-light datasets DroneRGBT and GAIIC2 to ensure that it improves the generalization of the model and avoids overfitting in a small number of single lights.

We utilize the Adam optimization algorithm with a fixed learning rate of \(10^{-4}\) to adjust the model parameters. Given that the ResNet50 backbone network weights are pretrained on ImageNet, we employ a smaller learning rate of \(10^{-5}\). The training is performed with a batch size of 4 for a total of 500 epochs. We conduct point proposal matching on the feature map with strides of 16, setting the number of reference points \(K\) to 4. This configuration is determined based on the dataset's statistical information to ensure that \(M > N\).Our point prediction mechanism employs shared prediction heads, which are composed of four layers with hidden layer dimensions of \([256, 512, 1024, 2048]\). For point regression, we set \(\gamma = 100\), and the weight parameter \(\tau\) for the matching process is configured as \(2 \times 10^{-2}\). The weight parameters \(\beta_1\), \(\beta_2\), \(\beta_3\), \(\beta_4\) in the loss function are set to 2.0, 2.0, 0.1, and 1.0 respectively. The parameters \(\gamma_1\) and \(\gamma_2\) are assigned values of 10 and 2. The weight parameters \(\lambda_1\), \(\lambda_2\), \(\lambda_3\), \(\lambda_4\) for auxiliary point matching are configured as 0.5, \(2 \times 10^{-4}\), \(2 \times 10^{-4}\), and 0.2 to balance the contributions of different components.All models are trained using the PyTorch framework on an NVIDIA A800 GPU.


\section{Experimental Results} 

In this section, the effectiveness of our proposed method is demonstrated by comparing it with state-of-the-art specialized architectures on standard benchmarks, and we also conduct a series of ablation experiments to evaluate our proposed strategy.

\subsection{Experimental} 

This section outlines our comparative analysis of population counting methods, in which our approach is benchmarked against a range of state-of-the-art techniques in different datasets. First, we evaluate the counting performance of our model on single-light images according to methods based on density map MMCount ~\cite{khan2024multimodal}, detection-based DroneNet ~\cite{wang2023dronenet} and point-based P2PNet ~\cite{song2021rethinking}, CLTR ~\cite{liang2022an}, PET ~\cite{liu2023pointquery}, APGCC ~\cite{chen2024improving}. Then, multimodal population counting using RGB-TIR images is used to evaluate the counting performance of our model on bimodal. Our experiments (see Table~\ref{table:sota} for details) highlight the leading performance of TAPNet, with the best results shown in bold and the next best results underlined.

The experimental results in Table~\ref{table:sota} focus on the DroneRGBT and GAIIC2 datasets, and the stability and accuracy of the TAPNet model for counting and localization on crowd counting tasks are demonstrated in the results of the evaluation metrics MAE, MSE, and F1-Score (results at threshold 0.8). In the first set of single-light RGB experiments, the point-based methods, except CLTR [18], TAPNet, P2PNet ~\cite{song2021rethinking}, PET ~\cite{liu2023pointquery}, and APGCC ~\cite{chen2024improving} significantly outperformed the density map MMCount ~\cite{khan2024multimodal} and the detection DroneNet ~\cite{wang2023dronenet} based methods, with which the point-based methods have greater potential. In the second set of bi-optical experiments, the bimodal fusion methods MMCount ~\cite{khan2024multimodal}, DroneNet ~\cite{wang2023dronenet}, and TAPNet show more excellent counting performance compared to the traditional methods based on single light, in which our method achieves a significant reduction in MAE and MSE, etc., even on the modal-misaligned GAIIC2 dataset, and the higher F1- Score further demonstrates the effectiveness of our method for accurate counting and localization in crowd counting tasks.

\subsection{Ablation studies} 

In this subsection, we will perform a series of ablation experiments to analyze the contribution made by the method proposed in this paper as well as the impact of the hyperparameters involved, only Table~\ref{table:spatial_offset} performs the ablation experiments on the GAIIC dataset, the rest of the experiments use the DroneRGBT dataset.

\vspace{-10pt}
\paragraph{Validity of the header point counting frame.} We implemented the interconversion between point labels and horizontal frame labels through a script, which was used to validate the significance of head point and frame counting. As shown in Table~\ref{table:Point}, the counting model for crowd head point detection decreases the MAE by 1.66 and improves the F1 score by 0.032 compared to the traditional counting model based on body frame detection.Therefore, the counting model for crowd head point detection is more counting advantageous and can further improve the robustness of the model for counting in densely occluded crowds.

\vspace{-10pt}
\paragraph{APG parameter settings.} Work ~\cite{song2021rethinking} has shown that the crowd counting framework based on auxiliary point guidance can effectively improve the stability of proposal-target matching. We verify the effect of the number of auxiliary positive and negative points on the performance of bi-optical crowd counting in Table~\ref{table:point_nums}, and the results show that using only auxiliary positive points for bi-optical data is more likely to utilize the model to select the optimal proposal, due to the fact that the fused bi-optical image has its features more distinct, making it easier to be selected. In addition, Table~\ref{table:point_random} shows that the exact auxiliary point random range is crucial for obtaining optimal results, as the densities of the datasets used in this paper are more balanced in comparison. Thus the experimental results are consistent with our analysis showing that relatively sparse data can be selected with smaller randomized auxiliary points, whereas an excessively large range of randomized auxiliary points is more suitable for dense data.

\vspace{-10pt}
\paragraph{Effectiveness of Dual-Light Fusion Modules.} By designing selectable fusion modules to adapt to different counting scenarios, we explored the impact of different fusion strategies with the following setups: (a) simple fusion of RGB and TIR images at the feature level only, (b) feature-level fusion using the dual-optical attentional fusion module, and (c) early fusion using the adaptive fusion module. The experimental results shown in Table~\ref{table:dual_fusion} indicate that while strategy (a) looks intuitive and has a straightforward design, it severely underestimates the accuracy of model localization and is therefore more suitable for simple and efficient counting tasks. (b) well solves the problem of inaccurate localization caused by systematic misalignment between image pairs, and localizes more accurately compared to (a), and Table~\ref{table:model_eva} further shows that (b) is capable of greater model counting and localization performance with the addition of fewer parameters.

\vspace{-10pt}
\paragraph{Validity of spatial random offsets.} As shown in Table~\ref{table:spatial_offset}, the spatial random offset data enhancement strategy reduces the MAE by 1.27 and 1.11 and improves the F1 scores by 0.047 and 0.035 on the two models compared to the method without the spatial random offset data enhancement strategy, which demonstrates that the spatial random offset data enhancement strategy is able to effectively improve the prediction accuracy and performance of the model.


\section{Conclusion}

In this paper, we propose the Two-Optical Attention Fusion Crowd Head Point Counting Model (TAPNet) to address the challenges in point-based crowd counting and localization tasks. To address the imaging limitations of a single sensor under adverse conditions such as nighttime, we propose the Attention Fusion Module (DAFP), which enhances modal fusion and interaction through complementary information from multimodal images to improve crowd counting performance. Aiming at the problem of inaccurate localization caused by systematic misalignment between image pairs, we also propose an adaptive two-branch feature decomposition fusion module (AFDF) in this paper. In addition, we employ a spatial random offset data enhancement strategy, which is used to further improve the generalization ability of the model. Extensive experimental results demonstrate that the approach in this paper exhibits excellent performance in crowd counting tasks by comparing it with a variety of state-of-the-art modeling architectures on benchmarks.

Despite the effectiveness of the proposed method, certain limitations still exist. For example, due to the image size, downsampling is not performed in the decomposition fusion stage and the image needs to be cropped to a smaller size for fusion. In future work, we will explore adaptive fusion methods to solve this problem.

{\small
\bibliographystyle{ieee_fullname}
\bibliography{egbib}
}

\end{document}